\pdfoutput=1
\documentclass[runningheads,orivec]{llncs}

\newif\ifdraft
\drafttrue
\draftfalse

\usepackage{times}
\usepackage{latexsym,amssymb,amsmath}
\usepackage{xspace}
\usepackage{comment}
\usepackage{url}
\usepackage{enumerate}
\usepackage[defblank]{paralist}
\usepackage{microtype}
\usepackage{hyperref}

 \usepackage[normalem]{ulem} % to override default set by changes package
 \ifdraft
   \usepackage{mychanges}
   \usepackage[colorinlistoftodos]{todonotes}
 \else
   \usepackage[final]{mychanges}
   \usepackage[disable]{todonotes}
 \fi
 \setauthormarkup[right]{{\scriptsize[#1]}}
 \definechangesauthor{DC}{orange}
 \definechangesauthor{MM}{green}
 \definechangesauthor{AS}{blue}
 \newcommand{\asd}[1]{\deleted[AS]{#1}} %ario deleted
 \newcommand{\asa}[1]{\added[AS]{#1}} %ario added
\newcommand{\ctx}{C}

\newcommand{\ctxb}{\mathit{cx}}
\newcommand{\initCtx}{\ctx_0}
\newcommand{\langCtx}{\L_{\ctxb}}
\newcommand{\ctxl}{\varphi_{\ctx}}

\newcommand{\cover}[1][d]{\prec_{#1}}
\newcommand{\cval}[2]{[#1\leadsto #2]}

\newcommand{\ctxAttSym}{\mathbb{C}_{dim}\xspace}
\newcommand{\ctxAttDom}[1]{Dom(#1)\xspace}
\newcommand{\cdom}[1][d]{\ensuremath{\mathit{Dom}(#1)}\xspace}

\newcommand{\ctxAtt}{context dimension\xspace}
\newcommand{\ctxAtts}{context dimensions\xspace}
\newcommand{\ctxAttV}{d}

\newcommand{\topdi}{\top_{\ctxAttV_i}}

\newcommand{\ctxThSym}{\Phi}
\newcommand{\ctxTh}{\ctxThSym_{\ctxAttSym}}

\newcommand{\ctxKB}{context-sensitive knowledge base\xspace}
\newcommand{\cKB}{CKB\xspace}
\newcommand{\cKBSym}{\O_\ctxb}

\newcommand{\contextualizedTBox}{contextualized TBox\xspace} %contextualized TBox
\newcommand{\cTBoxSym}{\T_\ctxb} %contextualized TBox

\newcommand{\tboxVoc}[1]{\textsc{voc}(#1)}

\newcommand{\tbox}{\T} % TBox
\newcommand{\initABox}{A_0} % Initial ABox
\newcommand{\actSet}{\Gamma} % action set

\newcommand{\act}{\alpha} % action set
\newcommand{\actParam}{p_1,\ldots,p_n}
\newcommand{\actForm}{\act(\actParam):\set{e_1,\ldots,e_m}} % action set
\newcommand{\effectSpec}{\map{[q^+_i] \land Q^-_i}{A_i}} % action set

\newcommand{\procSet}{\Pi} % set of process
\newcommand{\ctxProcSet}{\Pi_\ctx} %
\newcommand{\kab}{KAB\xspace}
\newcommand{\kabs}{KABs\xspace}
\newcommand{\kabSym}{\K}
\newcommand{\kabTup}{\tup{\tbox, \initABox, \actSet, \procSet}}

\newcommand{\cskab}{CKAB\xspace}
\newcommand{\cskabs}{CKABs\xspace}
\newcommand{\cskabSym}{\K_{\ctxb}}
\newcommand{\cskabTup}{\tup{\cTBoxSym, \initABox, \actSet, \procSet, \initCtx, \ctxProcSet}}

\newcommand{\const}{\Delta} %A set of constants
\newcommand{\servCall}{\F} %A set of function call
\newcommand{\map}[2]{#1 \rightsquigarrow #2} %for effect specifications
\newcommand{\adom}[1]{\textsc{adom}(#1)} %initial active domain
\newcommand{\carule}[2]{\tup{#1} \mapsto #2} %condition-action rules
\newcommand{\stateSet}{\Sigma} % set of states
\newcommand{\trans}{\Rightarrow} % set of transitions
\newcommand{\abox}{\mathit{abox}} % function to get the current abox
\newcommand{\cntx}{\mathit{ctx}} % function to get the current context

\newcommand{\TS}[1]{\Upsilon_{#1}} %transition system
\newcommand{\TSTup}{\tup{\const, \cTBoxSym, \stateSet, s_0, \abox, \cntx, \trans}} %transition system
\newcommand{\scmap}{\ensuremath{m}\xspace} % service call map
\newcommand{\scset}{\ensuremath{\mathbb{SC}}\xspace} % service call set

\newcommand{\doo}[1]{\textsc{do}(#1)} % do action exec
\newcommand{\calls}[1]{{\textsc{calls}({#1})}} % set of skolem terms
\newcommand{\eval}[1]{{\textsc{evals}(#1)}} % evaluation of service calls
\newcommand{\exec}[1]{\textsc{exec}_{#1}\xspace} %exec relation
\newcommand{\cexec}[1]{\textsc{cexec}_{#1}\xspace}

\newcommand{\midStateIndividuals}{\mathsf{inter}}
\newcommand{\midStateConcept}{\mathsf{State}}
\newcommand{\midState}{\midStateConcept(\midStateIndividuals)}

\newcommand{\kabTSTup}{\tup{\const,\tbox,\stateSet,s_0,\abox,\trans}} %transition system
\newcommand{\kabStateTup}{\tup{\A, \scmap}} %tuple state
\newcommand{\kabStateTupb}{\tup{\A', \scmap'}} %tuple state
\newcommand{\muL}{\mu\L} % MuL - modal mu calculus - language
\newcommand{\muladom}{\ensuremath{\muL_{A}^{{\textnormal{EQL}}}}\xspace}

\newcommand{\ctxMuL}{\textsc{ctx}}
\newcommand{\mulcs}{\ensuremath{\muL_{\ctxMuL}}\xspace}

\newcommand{\vfo}{\ensuremath{v}} %variable valuation
\newcommand{\vso}{\ensuremath{V}} %second order variable valuation

\newcommand{\MOD}[1]{(#1)^{\TS{}}}
\newcommand{\MODA}[1]{(#1)_{\vfo,\vso}^{\TS{}}}
\newcommand{\MODAX}[2]{(#1)_{\vfo #2,\vso}^{\TS{}}}

\newcommand{\dllitea}{\textit{DL-Lite}\ensuremath{_{\mathcal{A}}}\xspace}

\newcommand{\funct}[1]{(\ex{funct}~#1)}

\newcommand{\dlkbSym}{\O}
\newcommand{\dlkbTup}{\tup{\tbox, \A}}
\newcommand{\dlkb}{\dlkbSym = \dlkbTup}

\newcommand{\Ans}[2][]{\textsc{ans}_{#1}(#2)}
\newcommand{\ans}[2][]{\mathit{ans}_{#1}(#2)}

\newcommand{\conj}{\mathit{conj}}

\newcommand{\true}{\mathsf{true}}
\newcommand{\false}{\mathsf{false}}

\newcommand{\ex}[1]{\mathsf{#1}}

\def\qedboxempty{\vbox{\hrule\hbox{\vrule\kern3pt
\vbox{\kern3pt\kern3pt}\kern3pt\vrule}\hrule}}

\newcommand{\A}{A}

\newcommand{\E}{\mathcal{E}} \newcommand{\F}{\mathcal{F}}
 
\newcommand{\I}{\mathcal{I}} 
\newcommand{\K}{\mathcal{K}} \renewcommand{\L}{\mathcal{L}}
 
\renewcommand{\O}{\mathcal{O}}

\newcommand{\T}{T}

\newcommand{\ra}{\rightarrow}

\newcommand{\per}{\mbox{\bf .}}                  % period

\newcommand{\set}[1]{\{#1\}}                      % set
\newcommand{\tup}[1]{\langle #1\rangle}            % tuple

\newcommand{\myi}{\emph{(i)}\xspace}
\newcommand{\myii}{\emph{(ii)}\xspace}

\newcommand{\dom}[1][\I]{\Delta^{#1}}  % delta^I (domain of interpretation \I)
\newcommand{\Int}[2][\I]{#2^{#1}}      % #2^I    (interpretation function)
\newcommand{\NOT}{\neg}

\newcommand{\SOMET}[1]{\exists #1}

\newcommand{\INV}[1]{#1^{-}}

\newcommand{\limp}{\rightarrow}

\newcommand{\BOX}[1]{ [\!-\!] #1}
\newcommand{\DIAM}[1]{\langle \!-\! \rangle #1}

\newcommand{\ISA}{\sqsubseteq}

\newcommand{\exo}[1]{\ensuremath{\mathsf{\footnotesize #1}}\xspace} % ontology vocabulary
\newcommand{\exc}[1]{\texttt{\footnotesize #1}} %context vocabulary
\newcommand{\exv}[1]{\textsc{\footnotesize #1}} % context value vocabulary
\newcommand{\exm}[1]{\text{\footnotesize #1}} % exmath
\newcommand{\exx}[1]{\text{\footnotesize #1}} % others

\newcommand{\rws}{\rightsquigarrow}

\newcommand{\cov}[1]{\prec_{#1}}

\begin{document}

\title{Adding Context to Knowledge and Action Bases\thanks{This paper
is an abridged version of a paper published in the proceeding of JELIA~2014 \cite{JELIA-2014-kabs}.}}
\titlerunning{Adding Context to Knowledge and Action Bases}

\author{Diego Calvanese\inst{1} \and
 {\.I}smail {\.I}lkan Ceylan\inst{2} \and
 Marco Montali\inst{1} \and
 Ario Santoso\inst{1}}

\authorrunning{Diego Calvanese, {\.I}smail {\.I}lkan Ceylan, Marco Montali, and Ario Santoso}

\institute{Free University of Bozen-Bolzano,
 \email{\textit{lastname}@inf.unibz.it} \and
 Technische Universit\"at Dresden, \email{ceylan@tcs.inf.tu-dresden.de}}

\maketitle
%%
%% Proceedings Production
%%
\setcounter{page}{25}
\thispagestyle{plain}

\begin{abstract}
  Knowledge and Action Bases (KABs) have been recently proposed as a
  formal framework to capture the dynamics of systems which manipulate
  Description Logic (DL) Knowledge Bases (KBs) through action
  execution.  In this work, we enrich the KAB setting with contextual
  information, making use of different context dimensions.  On the one
  hand, context is determined by the environment using
  context-changing actions that make use of the current state of the
  KB and the current context.  On the other hand, it affects the set
  of TBox assertions that are relevant at each time point, and that
  have to be considered when processing queries posed over the
  KAB. Here we extend to our enriched setting the results on
  verification of rich temporal properties expressed in mu-calculus,
  which had been established for standard KABs.  Specifically, we show
  that under a run-boundedness condition, verification stays decidable\asd{
  and does not incur in any additional cost in terms of worst-case
  complexity}. 
\end{abstract}

%\endinput

%%% Local Variables:
%%% mode: latex
%%% TeX-master: "main"
%%% save-place: t
%%% End:

%\input{1-introduction}

\section{Introduction}
\label{sec:introduction}

Recent work in the areas of knowledge representation, databases, and
business processes \cite{Vian09,BCMD*13,CDLMS12,LDLHV12} has
identified the need for integrating static and dynamic aspects in the
design and maintenance of complex information systems.  The
\emph{static} aspects are characterized on the one hand by the data
manipulated by the system, and on the other hand by possibly complex
domain knowledge that may vary during the evolution of the system.
Instead, \emph{dynamic} aspects are affected by the processes that
operate over the system, by executing actions that manipulate the
state of the system.  In such a setting, in which new data may be
imported into the system from the outside environment, the system
becomes infinite-state in general, and the verification of temporal
properties becomes more challenging: indeed, neither finite-state
model checking \cite{ClGP99} nor most of the current techniques for
infinite-state model checking apply to this case.

\emph{Knowledge and action bases} (KABs) \cite{BCMD*13} have been
introduced recently as a mechanism for capturing systems in which
knowledge, data, and processes are combined and treated as first-class
citizens.  In particular, KABs provide a mechanism to represent
semantically rich information in terms of a description logic (DL)
knowledge base (KB) and a set of actions that manipulate such a KB
over time.  Additionally, actions allow one to import into the system
fresh values from the outside, via service calls.  In this setting,
the problem of verification of rich temporal properties expressed over
KABs in a first-order variant of the $\mu$-calculus has been studied.
Decidability has been established under the assumptions that in the
properties first-order quantification across states is restricted, and
that the system satisfies a so-called \emph{run-boundedness}
condition.  Intuitively, these ensure that along each run the system
cannot encounter (and hence manipulate) an unbounded number of
distinct objects.
In KABs, the intensional knowledge about the domain, expressed in
terms of a DL TBox, is assumed to be fixed along the evolution of the
system, i.e., independent of the actual state.  However, this
assumption is in general too restrictive, since specific knowledge
might hold or be applicable only in specific, \emph{context-dependent}
circumstances.  Ideally, one should be able to form statements that
are known to be true in certain cases, but not necessarily in all.

Work on representing and formally reasoning over contexts dates back
to work on generality in AI see \cite{Mcca87}. Since then, there has
been some effort in knowledge representation and in DLs to devise
context-sensitive formalisms, ranging from multi-context systems
\cite{BoSe03} to many-dimensional logics \cite{KlGu10}.  An important
aspect in modeling context is related to the choice of which kind of
information is considered to be fixed and which context dependent.
Specifically, for DLs, one can define the assertions in the TBox
\cite{BaKP12,CePe14}, the concepts \cite{BoSe03}, or both
\cite{SeHo12,KlGu10} as context-dependent.  Each choice addresses
different needs, and results in differences in the complexity of
reasoning.

We follow here the approach of \cite{BaKP12,CePe14}, and introduce
\emph{contextualized TBoxes}, in which each inclusion assertion is
adorned with context information that determines under which
circumstances the inclusion assertion is considered to hold.  The
relation among contexts is described by means of a lattice in
\cite{BaKP12} and by means of a directed acyclic graph in
\cite{CePe14}. In our case, we represent context using a finite set of
context dimensions, each characterized by a finite set of domain
values that are organized in a tree structure. If for a context
dimension $d$, a value $v_2$ is placed below $v_1$ in the tree (i.e.,
$v_2$ is a descendant of $v_1$), then the context associated to $v_1$
is considered to be more general than the one for $v_2$, and hence
whenever context dimension $d$ is in value $v_2$, it is also in value
$v_1$.

Starting from this representation of contexts, we enrich KABs towards
\emph{context-sensitive KABs} (\cskabs), by representing the
intensional information about the domain using a contextualized TBox,
in place of an ordinary one.  Moreover, the action component of KABs,
which specifies how the states of the system evolve, is extended in
\cskabs with \emph{context changing actions}.  Such actions determine
values for context dimensions in the new state, based on the data and
the context in the current state.  In addition, also regular
state-changing actions can query, besides the state, also the context,
and hence be enabled or disabled according to the context.
Notably, we show that verification of a very rich temporal logic,
which can be used to query the system evolution, contexts, and data,
is decidable for run-bounded \cskabs.
\asd{We also discuss how to recast the syntactic condition of
  \emph{weak acyclicity} \cite{BCMD*13}, which ensures
  run-boundedness, to the case of \cskabs.}

%\endinput

%%% Local Variables:
%%% mode: latex
%%% TeX-master: "main"
%%% save-place: t
%%% End:

%\input{2-premilinaries}

\section{Preliminaries}
\label{sec:prelim}

\smallskip
\noindent
\textbf{\dllitea.} For expressing knowledge bases, we use the
lightweight DL \dllitea \cite{CDLL*09}.  The syntax for \emph{concept}
and \emph{role} expressions in \dllitea is as follows:
\[
   \begin{array}{rcl@{}l}
     B &~::=~ &  N &~\mid~ \SOMET{R} \\
   \end{array}
   \qquad\qquad
   \begin{array}{rcl@{}l}
     R &~::=~ &  P &~\mid~ \INV{P}\\
   \end{array}
\]
where
\begin{inparaenum}[]
\item $N$ denotes a \emph{concept name},
\item $B$ a \emph{basic concept},
\item $P$ a \emph{role name},
\item $\INV{P}$ an \emph{inverse role}, and
\item $R$ a \emph{basic role}.
\end{inparaenum}
A \dllitea \emph{knowledge base} (KB) is a tuple $\dlkb$, where:
\begin{compactitem}
\item $\tbox$ is a TBox, containing a finite set of assertion of the form:
\[
  B_1 \ISA B_2 \qquad\quad
  R_1 \ISA R_2 \qquad\quad
  B_1 \ISA \neg B_2 \qquad\quad
  R_1 \ISA \NOT R_2\qquad\quad
  \funct{R}
\]
From left to right, assertions of the first two columns respectively
denote \emph{positive inclusions} between basic concepts and basic
roles; assertions of the third and fourth columns denote
\emph{negative inclusions} between basic concepts and basic roles;
assertions of the last column denote \emph{functionality} on roles.
\item $\A$ is an Abox, i.e., a finite set of \emph{ABox membership assertions}
  of the form $ N(c_1)$ or $P(c_1,c_2)$, where $c_1$, $c_2$ denote individuals
  (constants).
\end{compactitem}
We use the standard semantics of DLs based on FOL interpretations
$\I=(\dom,\Int{\cdot})$ such that $\Int{c}\in\dom$, $\Int{N}\subseteq\dom$, and
$\Int{P}\subseteq\dom\times\dom$.  The semantics of the \dllitea constructs and
of TBox and ABox assertions, and the notions of \emph{satisfaction} and of
\emph{model} are as usual (see, e.g., \cite{CDLLR07}).
We also say that $\A$ is \emph{$\tbox$-consistent} if $\dlkb$ is
satisfiable, i.e., admits at least one model.
% otherwise we say $A$ is \emph{$T$-inconsistent}.

\smallskip
\noindent
\textbf{Queries.}
We are interested to query the KB, i.e., retrieving relevant constants
in the ABox based on the query. We denote with $\adom{\A}$ the \emph{set of
constants appearing in $\A$}.
A \emph{union of conjunctive queries} (UCQ) $q$ over a KB $\dlkb$ is a FOL
formula of the form
$\bigvee_{1\leq i\leq
  n}\exists\vec{y_i}\per\conj_i(\vec{x},\vec{y_i})$ with free
variables $\vec{x}$ and existentially quantified variables
$\vec{y}_1,\ldots,\vec{y}_n$.
Each $\conj_i(\vec{x},\vec{y_i})$ in $q$ is a conjunction of atoms of
the form $N(z)$, $P(z,z')$, where $N$ and $P$ respectively denote a
concept and a role name occurring in $\tbox$, and $z$, $z'$ are
constants in $\adom{\A}$ or variables in $\vec{x}$ or $\vec{y_i}$, for
some $1 \leq i \leq n$.

%--- certain answers
The \emph{(certain) answers} of $q$ over $\dlkb$ are defined as
the set $\ans{q,T,A}$ of substitutions $\sigma$ which substitute the
free variables of $q$ with constants from $\adom{A}$ such that
$q\sigma$ evaluates to true in every model of $\dlkb$.
If $q$ has no free variables, then it is called \emph{boolean} and its
certain answers are either $\true$ or $\false$.

%--- ECQ
We also consider an extension of UCQs, namely
\textit{EQL-Lite}(UCQ)~\cite{CDLLR07b} (briefly, ECQs), i.e., the FOL
query language whose atoms are UCQs evaluated according to the certain
answer semantics.
An \emph{ECQ} over a TBox $\tbox$ is a possibly open formula of the
form:
\[
  Q ~::=~  [q]  ~\mid~ \lnot Q ~\mid~ Q_1\land Q_2 ~\mid~
              \exists x\per Q
\]
%}
where $q$ is a UCQ over $\tbox$.
% and $[q]$ denotes that $q$ is evaluated under the (minimal)
% knowledge operator (cf.~\cite{CDLLR07b}).
%
The \emph{certain answers $\Ans{Q,\tbox,\A}$ of an ECQ $Q$ over $\dlkb$} are
obtained by first computing the certain answers over $\dlkb$ of each UCQs
embedded in $Q$, then evaluating them through the first-order part of $Q$, and
interpreting existential variables as ranging over $\adom{A}$.
%
% Hence, also computing $\Ans{Q,\tbox,\A}$ of an ECQ $Q$ over a
% \dllitea ontology $(\tbox,\A)$ is in $\ACzero$ in the size of $\A$
% \cite{CDLLR07b}.
%
As stated in \cite{CDLLR07b}, the reformulation algorithm for answering query
$q$ over \dllitea KB $\dlkb$ which allows us to ``compile away'' the TBox
(i.e., $\ans{q,T,A} = \ans{rew(q),\emptyset,A}$, where $rew(q)$ is a UCQ
computed by the algorithm in \cite{CDLL*09}) can be extended to ECQs.

%\subsection{Knowledge and Action Bases}
\smallskip
\noindent
\textbf{Knowledge and Action Bases.} In the following, we make use of
a countably infinite set $\const{}$ of \textit{constants}, and a
finite set $\servCall$ of \textit{functions} representing
\textit{service calls}, which can be used to introduce fresh values
from $\const{}$ into the system.

A \emph{knowledge and action base} (\kab) is a tuple $\kabSym = \kabTup$ where:
\begin{inparaenum}[\it(i)]
\item $\tbox$ is a \dllitea TBox capturing the domain of interest,
\item $\initABox$ is the initial \dllitea ABox, which intuitively represents
  the initial data of the system,
\item $\actSet$ is a finite set of actions that characterize the evolution of
  the system,
\item $\procSet$ is a finite set of condition-action rules forming a process
  that intuitively specifies when and how an action can be executed.
\end{inparaenum}
$\tbox$ and $\initABox$ together form the \emph{knowledge base} while $\actSet$
and $\procSet$ form the \emph{action base}.

%%-------- ACTION -------%%
An \textit{action} $\act \in \actSet$ represents the progression
mechanism that changes the ABox in the current state and hence
generates a new ABox for the successor state. Formally, an action
$\act \in \actSet$ is represented as $\actForm$ where
\begin{inparaenum}[\it (i)]
\item $\act$ is the \emph{action name},
\item $p_1,\ldots,p_n$ are the \emph{input parameters}, and
\item $\set{e_1,\ldots,e_m}$ is the set of \emph{effects}. Each effect $e_i$ is
  of the form $\effectSpec$, where:
  \begin{inparaenum}[(a)]
  \item $q^+_i$ is an UCQ, and $Q^-_i$ is an arbitrary ECQ whose free variables
    occur all among the free variables of $q^+_i$.
    % \footnote{The ECQ(UCQ) division is a convenience to have readily
    % available the positive part of the condition.}
    %
  \item $A_i$ is a set of facts (over the alphabet of $T$) which includes as
    terms: constants in $\adom{A_0}$, input parameters, free variables of
    $q^+_i$, and Skolem terms representing service calls formed by applying a
    function $f \in \servCall$ to one of the previous kinds of terms.
  \end{inparaenum}
  Intuitively, $q^+_i$, together with $Q^-_i$ acting as a filter, selects the
  values that instantiate the facts listed in $A_i$.  Collectively, the
  instantiated facts produced from all the effects of $\act$ constitute the
  newly generated ABox, once the ground service calls are substituted with
  corresponding results.
\end{inparaenum}
%%-------- END OF ACTION -------%%
%
%%-------- PROCESS -------%%
The \emph{process} $\procSet$ is formally defined as a finite set of
\emph{condition-action rules} of the form $Q(\vec{x})\mapsto\alpha(\vec{x})$,
where:
\begin{inparaenum}[\it (i)]
\item $\act\in\actSet$ is an action, and
\item $Q(\vec{x})$ is an ECQ over $\tbox$, which has the parameters of $\act$
  as free variables $\vec{x}$, and quantified variables or values in
  $\adom{\initABox}$ as additional terms.
\end{inparaenum}
%%-------- END OF PROCESS -------%%

\asd{
Notice that KABs are a pristine action specification framework, aimed at
understanding the interaction between the static and dynamic components of
systems evolving over time, towards general decidability results for
verification.  On top of KABs, several abstractions typical of reasoning about
actions in AI can be built, see, e.g., \cite{MoCD14}.
}

\smallskip
\noindent
\textbf{KABs Execution Semantics.}~ The execution semantics of a \kab is
defined in terms of a possibly infinite-state transition system.  Formally,
given a \kab $\kabSym = \kabTup$, we define its semantics by the
\emph{transition system} $\TS{\kabSym} = \kabTSTup$, where:
\begin{inparaenum}[\it (i)]
\item $\tbox$ is a \dllitea TBox;
\item $\stateSet$ is a (possibly infinite) set of states;
\item $s_0 \in \stateSet$ is the initial state;
\item $\abox$ is a function that, given a state $s\in\stateSet$,
  returns an ABox associated to $s$;
\item ${\Rightarrow} \subseteq \Sigma\times\Sigma$ is a transition
  relation between pairs of states.
\end{inparaenum}
Intuitively, the transitions system $\TS{\kabSym}$ of \kab $\kabSym$ captures
all possible evolutions of the system by the actions in accordance with the
process rules.
%
% Each states in $\TS{\kabSym}$ are characterized by the knowledge
% bases. Technically, the construction of $\TS{\kabSym}$ is starting
% from the initial ABox $\initABox$ by applying all possible
% executable action (manipulating the ABox) and we obtain all possible
% successor states, and then repeating this step forever.

%%--- Service call semantics - deterministic
During the execution, an action can issue service calls.  In this paper, we
assume that the semantics of service calls is \emph{deterministic}, i.e., along
a run of the system, whenever a service is called with the same input
parameters, it will return the same value.
To enforce this semantics, the transition system remembers the results of
previous service calls in a so-called service call map that is part of the
system state.  Formally, a \emph{service call map} is defined as a partial
function $\scmap:\scset\ra\const$, where $\scset$ is the set
$\{f(v_1,\ldots,v_n) \mid f/n \in \servCall \textrm{ and } \{v_1,\ldots,v_n\}
\subseteq \const \}$ of (skolem terms representing) \emph{service calls}.
%%--- END OF Service call semantics - deterministic
%
%%--- State formalization
Each state $s \in \stateSet$ of the transition system $\TS{\kabSym}$ is a tuple
$\kabStateTup$, where $\A$ is an ABox and \scmap is a service call map.
%%--- END OF State formalization

%%--- Action Execution Semantics
The semantics of an \emph{action execution} is as follows:
Given a state $s = \kabStateTup$,
let $\act \in \actSet$ be an action of the form $\actForm$ with $e_i =
\effectSpec$, and let $\sigma$ be a \emph{parameter substitution} for
$\actParam$ with values taken from $\const$.
%
%We say that $\sigma$ is \emph{legal parameter} for $\act$ in state
%$s$ w.r.t. condition-action rule $\carule{Q,
%  \actCtx}{\act} \in \procSet$ if  $\tup{p_1,\ldots,p_r}\sigma \in
%\Ans{Q,\A}$.
%
We say that \emph{$\act$ is executable in state $s$ with parameter
  substitution $\sigma$}, if there exists a condition-action rule
$Q(\vec{x})\mapsto\alpha(\vec{x}) \in \procSet$
% and parameter substitution $\sigma$
s.t.\ $\Ans{Q\sigma, \tbox,\A}$ is $\true$.
% In that case we call $\sigma$ a \emph{legal} parameter substitution for
% $\act$.%
The result of the application of $\act$ to an ABox $\A$ using a parameter
substitution $\sigma$ is captured by the following function:
%{%\small
%\vspace*{-1mm}
\[
  \doo{\tbox, \A, \act\sigma} = \bigcup_{\effectSpec \text{ in } \act\ }
  \bigcup_{\rho\in\Ans[]{([q_i^+]\land Q_i^-)\sigma,\tbox,\A}} A_i\sigma\rho
%\vspace*{-1mm}
\]
%}
Intuitively, the result of the evaluation of $\alpha$ is obtained by combining
the contribution of each effect of $\alpha$, which in turn is obtained by
grounding the facts $A_i$ in the head of the effect with all the certain
answers of the query $[q_i^+]\land Q_i^-$ over $\tup{T,A}$.
%%--- END OF Action Execution Semantics

%%--- Service call evaluation
The result of $\doo{\tbox, \A, \act\sigma}$ is in general not a proper ABox,
because it could contain (ground) Skolem terms, attesting that in order to
produce the ABox, some service calls have to be issued.  We denote by
$\calls{\doo{\tbox, \A, \act\sigma}}$ the set of such ground service calls, and
by $\eval{\tbox,\A,\act\sigma}$ the set of substitutions that replace such
calls with concrete values taken from $\const$. Specifically,
$\eval{\tbox,\A,\act\sigma}$ is defined as
%
%{%\small
%\vspace*{-1.0mm}
\[
  \eval{\tbox,\A,\act\sigma} = \{ \theta\ \mid\ \theta:
  \calls{\doo{\tbox, \A, \act\sigma}} \ra \const
  \text{ is a total function} \}.
%\vspace*{-4.5mm}
\]
%}
%%--- END OF Service call evaluation

%%--- EXEC Transition relation
With all these notions in place, we can now recall the execution
semantics of a \kab $\kabSym = \kabTup$. To do so, we first introduce
a transition relation  $\exec{\kabSym}$
that connects pairs of ABoxes and service call maps due to action
execution. In particular, $\tup{\kabStateTup,\act\sigma,\kabStateTupb}
\in \exec{\kabSym}$ if the following holds:
\begin{inparaenum}[\it (i)]
\item $\act$ is \emph{executable} in state $s = \kabStateTup$ with parameter
  substitution $\sigma$;
\item there exists $\theta \in \eval{\tbox,\A,\act\sigma}$ s.t. $\theta$ and $\scmap$
  ``agree'' on the common values in their domains (in order to realize the
  deterministic service call semantics);
\item $\A' = \doo{\tbox, \A, \act\sigma}\theta$;
\item $\scmap' = \scmap \cup \theta$ (i.e., updating the history of
  issued service calls).
\end{inparaenum}
%% --- END OF EXEC Transition relation

%-------- DEFINE THE CS-TS
The transition system $\TS{\kabSym}$ of $\kabSym$ is then defined as
$\kabTSTup$ where
\begin{inparaitem}[]
\item $s_0 = \tup{\initABox,\emptyset}$, and
\item $\stateSet$ and $\trans$ are defined by simultaneous induction as the
  smallest sets satisfying the following properties:
  \begin{inparaenum}[\it (i)]
  \item $s_0 \in \stateSet$;
  \item if $\tup{\A,\scmap} \in \stateSet$, then for all actions $\act \in
    \actSet$,
    for all substitutions $\sigma$ for the parameters of $\act$ and
    for all $\tup{\A',\scmap'}$ s.t.
    $\tup{\tup{\A,\scmap}, \act\sigma,\tup{\A',\scmap'}} \in \exec{\kabSym}$
    and $\A'$ is $\tbox$-consistent,
    we have $\tup{\A',\scmap'}\in\stateSet$, $\tup{\A,\scmap}\trans
    \tup{\A',\scmap'}$.
  \end{inparaenum}
\end{inparaitem}
A \emph{run} of $\TS{\kabSym}$ is a (possibly infinite) sequence $s_0s_1\cdots$
of states of $\TS{\kabSym}$ such that $s_i\trans s_{i+1}$, for all $i\geq 0$.

%\endinput

%%% Local Variables:
%%% mode: latex
%%% TeX-master: "main"
%%% save-place: t
%%% End:

%\input{3-ContextualizedKB}

\section{Contextualizing Knowledge Bases}
\label{sec:ContextualizingTheKnowledgeBases}

Following \cite{McCa93}, we formalize context as a mathematical object.
Basically, we follow the approach in \cite{SeHo12} of contextualizing
knowledge bases by adopting the metaphor of considering context as a box
\cite{BoGS13,GiBo97}.  Specifically, this means that the knowledge
represented by the TBox (together with the ABox) in a certain context is
affected by the values of parameters used to characterize the context itself.

Formally, to define the context, we fix a set of variables $\ctxAttSym =
\{d_1,\ldots,d_n\}$ called \emph{\ctxAtts}. Each \ctxAtt $d_i \in \ctxAttSym$
comes with its own tree-shaped finite \emph{value domain}
$\tup{\cdom[d_i],\cover[d_i]}$, where $\cdom[d_i]$ represents the finite set of
domain values, and $\cover[d_i]$ represents the predecessor relation forming
the tree. We denote the domain value in the root of the tree with $\topdi$.
Intuitively, $\topdi$ is the most general value in the tree-shaped value
hierarchy of $\cdom[d_i]$.
We denote the fact that a \ctxAtt $d$ is in value $v$ by $\cval{d}{v}$, and
call this a \emph{\ctxAtt assignment}.

A \emph{context} $\ctx$ over a set $\ctxAttSym$ of \ctxAtts is defined as a set
$\{\cval{d_1}{v_1},\ldots,\cval{d_n}{v_n}\}$
of \ctxAtt assignments such that for each \ctxAtt $d\in\ctxAttSym$, there
exists exactly one assignment $\cval{d}{v}\in\ctx$.
To predicate over contexts, we introduce a \emph{context expression language}
$\langCtx$ over $\ctxAttSym$, which corresponds to propositional logic where
the propositional letters are \ctxAtt assignments over $\ctxAttSym$.  The
syntax of $\langCtx$ is as follows:
\[
  \ctxl ~::=~ \cval{d}{v} ~\mid~ \ctxl\land\ctxl' ~\mid~ \lnot\ctxl
\]
where $d\in\ctxAttSym$, and $v \in \ctxAttDom{d}$.  We adopt the standard
propositional logic semantics and the usual abbreviations.  The notion of
\emph{satisfiability} and \emph{model} are as usual. We call a formula
expressed in $\langCtx$ a \emph{context expression}.

Observe that a context $\ctx=\{\cval{d_1}{v_1},\ldots,\cval{d_n}{v_n}\}$, being
a set of (atomic) formulas in $\langCtx$, can be considered as a propositional
theory.
The semantics of value domains in $\ctxAttSym$ can also be characterized by a
$\langCtx$ theory.  Specifically, we define the theory $\ctxTh$ as the smallest
set of context expressions satisfying the following conditions. For every
\ctxAtt $d\in\ctxAttSym$, we have:
\begin{compactitem}%[\it (i)]
\item For all values $v_1,v_2\in\ctxAttDom{d}$ s.t.\ $v_1\cover v_2$, we have
  that $\Phi_{\ctxAttSym}$ contains the expression
  $\cval{d}{v_1}\limp\cval{d}{v_2}$.  Intuitively, this states that the
  value $v_2$ is more general than $v_1$, and hence, whenever we have
  $\cval{d}{v_1}$ we can infer that $\cval{d}{v_2}$.
\item For all values $v_1,v_2,v\in\ctxAttDom{d}$ s.t.\ $v_1\cover v$ and
  $v_2\cover v$, we have that $\Phi_{\ctxAttSym}$ contains the expression
  $\cval{d}{v_1}\limp\lnot\cval{d}{v_2}$.  Intuitively, this expresses that
  sibling values $v_1$ and $v_2$ are disjoint.
\end{compactitem}

\begin{example}
\label{ex:context}
\small
  Consider an online retail enterprise (e.g., amazon.com) with many
  warehouses. A simple order processing scenario is as follows:
\begin{inparaenum}[(i)]
\item The customer submits the order.
\item The central processing office receives the order.
\item The \emph{assembler} collects the ordered product. For each
  product that is not available in the central warehouse, the
  assembler makes a request to one of the warehouses having that product.
\item The \emph{wrapper} wraps the ordered product.
\item The \emph{quality controller (QC)} checks the prepared order.
\item The \emph{delivery team} delivers the order to the delivery service.
\end{inparaenum}
In this scenario we consider
$\ctxAttSym = \{ \exc{PP} , \exc{S}\}$, %, \exc{OA}\}$,
where \exc{PP} stands for \emph{processing plan}, and \exc{S} stands
for \emph{season}.
$\ctxAttDom{\exc{PP}} = \{\exv{WE},
\exv{ME}, \exv{RE}, \exv{N}, \exv{AP}\}$
(\begin{inparaenum}[]
\item \exv{WE} stands for \emph{worker efficiency},
\item \exv{ME} stands for \emph{material efficiency},
\item\exv{RE} stands for \emph{resource efficiency},
\item \exv{N} stands for \emph{normal processing plan}, and
\item\exv{AP} stands for \emph{any processing plan}.
\end{inparaenum}),
where
\begin{inparaenum}[(i)]
\item $\exv{WE} \cov{\exc{PP}}  \exv{RE} $,
\item $\exv{ME} \cov{\exc{PP}} \exv{RE} $,
\item $\exv{RE} \cov{\exc{PP}} \exv{AP} $,
\item $\exv{N} \cov{\exc{PP}} \exv{AP} $,
\end{inparaenum}
For example, $\exv{WE} \cov{\exc{PP}} \exv{RE} $ means that
\emph{worker efficiency} is a form of \emph{resource efficiency}.
%
%
% --- DOM S
$\ctxAttDom{\exc{S}} = \{\exv{WH}, \exv{PS}, \exv{LS}, \exv{NS}, \exv{AS}\}$
(\begin{inparaenum}[]
\item \exc{WH} stands for \emph{winter holiday},
\item \exc{PS} stands for \emph{peak season},
\item \exc{LS} stands for \emph{low season},
\item\exc{NS} stands for \emph{normal season}, and
\item\exc{AS} stands for \emph{any season}.
\end{inparaenum}),
where
\begin{inparaenum}[(i)]
\item $\exv{WH} \cov{\exc{S}}  \exv{PS} $,
\item $\exv{PS} \cov{\exc{S}} \exv{AS} $,
\item $\exv{NS} \cov{\exc{S}} \exv{AS} $,
\item $\exv{LS} \cov{\exc{S}} \exv{AS} $.
\end{inparaenum}
\end{example}

%-------------------------------------
% Context-sensitive knowledge bases (CKB)
%-------------------------------------
\smallskip
\noindent
\textbf{Context-Sensitive Knowledge Bases.}~ We define a \emph{\ctxKB} (\cKB)
$\cKBSym$ over $\ctxAttSym$ as a standard DL knowledge base in which the TBox
assertions are contextualized.  Formally, a \emph{contextualized TBox}
$\cTBoxSym$ over $\ctxAttSym$ is a finite set of assertions of the form
$\tup{t:\varphi}$, where $t$ is a TBox assertion and $\varphi$ is a context
expression over $\ctxAttSym$.  Intuitively, $\tup{t:\varphi}$ expresses that
the TBox assertion $t$ holds in all those contexts satisfying $\varphi$, taking
into account the theory $\Phi_{\ctxAttSym}$.
Given a contextualized TBox $\cTBoxSym$, we denote with $\tboxVoc{\cTBoxSym}$
the set of all concept and role names appearing in $\cTBoxSym$, independently
from the context.

Given a \cKB $\cKBSym=\tup{\cTBoxSym,\A}$ and a context $\ctx$, both over
$\ctxAttSym$, we define the \emph{KB $\cKBSym$ in context $\ctx$} as the KB
$\cKBSym^{\ctx}=\tup{\cTBoxSym^{\ctx},\A}$, where
$\cTBoxSym^{\ctx} =\{t\mid \tup{t:\varphi} \in \cTBoxSym \text{ and }
\ctx \cup \ctxTh \models \varphi\}$.

\begin{example}
  \small
  Continuing our example, in a normal situation, to guarantee a suitable
  service quality, \emph{wrapper} and \emph{assembler} must not be the
  \emph{QC}. However, in the situation (context) where we have either
  \emph{peak season} ($\cval{\exc{S}}{\exv{PS}}$) or the company wants to
  promote \emph{worker efficiency} ($\cval{\exc{PP}}{ \exv{WE}}$), the
  \emph{wrapper} and the \emph{assembler} act also as \emph{QC}.  This
  situation can be encoded as follows:
\[
\begin{array}[t]{ll}
\tup{\exo{Assembler} \sqsubseteq \neg \exo{QC}:
  \cval{\exc{PP}}{\exv{N}} \land \cval{\exc{S}}{\exv{NS}}}  \ \
&\tup{\exo{Assembler} \sqsubseteq \exo{QC}:
  \cval{\exc{PP}}{\exv{WE}} \lor \cval{\exc{S}}{\exv{PS}}}\\
\tup{\exo{Wrapper} \sqsubseteq \neg \exo{QC}:
  \cval{\exc{PP}}{\exv{N}} \land \cval{\exc{S}}{\exv{NS}}}
&\tup{\exo{Wrapper} \sqsubseteq \exo{QC}:
  \cval{\exc{PP}}{\exv{WE}} \lor \cval{\exc{S}}{\exv{PS}}}
\end{array}
\]
\end{example}

%\endinput

%%% Local Variables:
%%% mode: latex
%%% TeX-master: "main"
%%% save-place: t
%%% End:

%\input{4-CSKAB}

\section{Context-Sensitive Knowledge and Action Bases}
\label{sec:CSKAB}

We now enhance KABs with context-related information, introducing in
particular \emph{context-sensitive knowledge and action bases}
(\cskabs), which consist of:
\begin{inparaenum}[\it (i)]
\item a \ctxKB (\cKB), which maintains the information of interest,
\item an action base, which characterizes the system evolution, and
\item context information that evolves over time, capturing changing
  circumstances.
\end{inparaenum}
Differently from \kabs, where the TBox is fixed a-priori and remains
rigid during the evolution of the system, in \cskabs the TBox changes
depending on the current context.  Alongside the evolution mechanism
for data borrowed from KABs, \cskabs include also a progression
mechanism for the context itself, giving raise to a system in which
data and context evolve simultaneously.

\subsection{Formalization of \cskabs}
\label{sec:CSKABFormalization}

As for standard KABs, in addition to $\const{}$ and $\servCall$,  
we fix the set $\ctxAttSym = \{d_1, \ldots, d_n\}$ of \emph{\ctxAtts.}
A \cskab is a tuple $\cskabSym=\cskabTup$ where:
%\begin{inparaenum}[\it(i)]
\begin{compactitem}
\item $\cTBoxSym$ is a \dllitea \emph{\contextualizedTBox} capturing the
  domain of interest.
\item $\initABox$ and $\actSet$ are as in a KAB.
\item $\procSet$ is a finite set of condition-action rules that extend
  those of KABs by including, in the precondition, a context
  expression. Such context expression implicitly selects those
  contexts in which the corresponding action can be executed.
Specifically, each
  condition-action rule has the form
$\tup{Q(\vec{x}), \ctxl} \mapsto \alpha(\vec{x})$,
where
\begin{inparaenum}[\it (i)]
\item $\alpha \in \actSet$ is an action,
\item $Q(\vec{x})$ is an ECQ over $\cTBoxSym$ whose free variables $\vec{x}$
  correspond exactly to the parameters of $\alpha$, and
\item $\ctxl$ is a context expression over $\ctxAttSym$.
\end{inparaenum}
\item $\initCtx$ is the initial context over $\ctxAttSym$.
\item $\ctxProcSet$ is a finite set of context-evolution rules, each
  of which determines the configuration of the new context
  depending on the current context and data. Each
  \emph{context-evolution rule} has the form
$\tup{Q, \ctxl} \mapsto C_{new}$,
where:
\begin{inparaenum}[\it (i)]
\item $Q$ is a boolean ECQ  over $\cTBoxSym$,
\item $\ctxl$ is a context expression, and
\item $C_{new}$ is a finite set of \ctxAtt assignments such that for each \ctxAtt $d\in\ctxAttSym$, there
exists \emph{at most one} \ctxAtt assignment $\cval{d}{v}\in\ctx$. If
a context variable is not assigned by $C_{new}$, it maintains the
assignment of the previous state.
\end{inparaenum}
\end{compactitem}

\begin{example}
\small
In our running example, suppose the company has \emph{warehouses} in a remote
area (\emph{remote warehouses}), each of which is expected to guarantee a
certain \emph{time to delivery} (TTD) for products.  During the \emph{low
 season}, the company is free to set the TTD for all its remote warehouses,
which we model as a $\exo{chgTTD}()$ action.  The execution of this action is
controlled by the condition-action rule $ \carule{\exo{\exists w. RemWH(w)},
 \cval{\exc{S}}{\exv{LS}}}{\exo{chgTTD}()}.  $ Assuming that the company
maintains the TTD for a remote warehouse in the relation $\exo{hasTTD}$, the
$\exo{chgTTD}()$ action can be specified as follows:
%\[
$  \exo{chgTTD}():\{~
  \begin{array}[t]{@{}l}
    \exo{RemWH}(x) \wedge \exo{hasTTD}(x, y)
    \rws \{\exo{RemWH}(x), \exo{hasTTD}(x, \exx{newTTD}(x,y))  \}
    \}
  \end{array}
$
%\]
Intuitively, the unique effect in $\exo{hasTTD}$ updates the TTD of a remote
warehouse $x$, by issuing a service call $\exx{newTTD}(x,y)$, which also takes
into account the current TTD $y$ of $x$.
\end{example}

\begin{example}
\small
An example of context-evolution rule is $\carule{\exm{true},
 \cval{\exc{S}}{\exv{PS}} }{\cval{\exc{S}}{\exv{NS}}}$. It models the
transition from \emph{peak season} to \emph{normal season}, independently from
the data.
\end{example}

\subsection{\cskab Execution Semantics}
\label{sec:CSKABExecutionSemantics}

We are interested in verifying temporal properties over the evolution of
\cskabs, in particular ``robust'' properties that the system is required to
guarantee independently from context changes.  Towards this goal, we define the
execution semantics of \cskabs in terms of a possibly infinite-state transition
system that simultaneously captures all possible evolutions of the system as
well as all possible context changes.

Each state in the execution of a \cskab is a tuple $\tup{id,\A, \scmap, C}$,
where $id$ is a state identifier, $A$ is an ABox maintaining the current data,
$m$ is a service call map accounting for the service call results obtained so
far, and $C$ is the current context. The context univocally selects which are
the axioms of the contextual TBox that currently hold, in turn determining the
current KB.

Formally, given a \cskab $\cskabSym = \cskabTup$, we define its semantics in
terms of a \emph{context-sensitive transition system} $\TS{\cskabSym} =\TSTup$,
where:
\begin{inparaenum}[\it (i)]
\item $\cTBoxSym$ is a \contextualizedTBox;
\item $\stateSet$ is a set of states;
\item $s_0 \in \stateSet$ is the initial state;
\item $\abox$ is a function that, given a state $s\in\stateSet$,
  returns the ABox associated to $s$;
\item $\cntx$ is a function that, given a state $s\in\stateSet$,
  returns the context associated to $s$;
\item ${\Rightarrow} \subseteq \Sigma\times\Sigma$ is a transition
  relation between pairs of states.
\end{inparaenum}

Starting from the initial state $s_0$, $\TS{\cskabSym}$ accounts for all the
possible (simultaneous) data and context transitions. To single out the
dynamics of the system as opposed to those of the context, the transition
system is built by repeatedly alternating between system and context
transitions.
Technically, we revise the notion of executability for KABs by taking into
account context expressions, as well as the context evolution.  Given an action
$\alpha \in \actSet$, we say that $\act$ is \emph{executable} in state $s$ with
parameter substitution $\sigma$ if there exists a condition-action rule
$\tup{Q(\vec{x}),\ctxl} \mapsto \alpha(\vec{x})$ in $\procSet$
% and parameter substitution $\sigma$
s.t.\
$\vec{x}\sigma \in \Ans{Q, \cTBoxSym^{\cntx(s)},\abox(s)}$ and
$\cntx(s) \cup \ctxTh \models \ctxl$.

%%--- EXEC Transition relation
We then introduce an \emph{action transition relation} $\exec{\cskabSym}$,
where $\tup{\tup{A,\scmap,C},\act\sigma,\tup{A',\scmap',C'}} \in
\exec{\cskabSym}$ if the following holds:
\begin{compactitem}
\item Action $\act$ is \emph{executable} in state $\tup{A,\scmap,C}$ with
  parameter substitution $\sigma$;
\item There exists $\theta \in \eval{\cTBoxSym^C,A,\act\sigma}$ s.t.\ $\theta$
  and $\scmap$ ``agree'' on the common values in their domains;
\item $\A' = \doo{\cTBoxSym^\ctx, \A, \act\sigma}\theta$;
\item $\scmap' = \scmap \cup \theta$;
\item $C' = C$, i.e., the context does not change.
\end{compactitem}
Alongside the action transition relation, we also define a \emph{context
 transition relation} $\cexec{\cskabSym}$, where
$\tup{\tup{A,\scmap,C},\tup{A',\scmap',C'}} \in \cexec{\cskabSym}$ if the
following holds:
\begin{compactitem}
\item $A' = A$, i.e., the ABox does not change;
\item $\scmap' = \scmap$, i.e., the service call map does not change;
\item there exists a context rule $\tup{Q,\ctxl} \mapsto C_{new}$ in
  $\ctxProcSet$ s.t.:
\begin{inparaenum}[\it (i)]
\item $\Ans{Q, \cTBoxSym^C,A}$ is $\true$;
\item $C \cup \ctxTh \models \ctxl$;
\item for every context dimension $d\in\ctxAttSym$ s.t.\ $\cval{d}{v}
  \in C_{new}$, we have $\cval{d}{v} \in C'$;
\item for every context dimension $d\in\ctxAttSym$ s.t.\ $\cval{d}{v}
  \in C$, and there does not exist any $v_2$ s.t.\ $\cval{d}{v_2} \in
  C_{new}$, we have $\cval{d}{v} \in C'$.
\end{inparaenum}
\end{compactitem}
Given these, we can now define how $\TS{\cskabSym}$ is constructed, by suitably
alternating the action and context transitions. In order to single out the
states obtained by applying just an action transition and for which the context
transition has not taken place yet,
% whether a state is obtained by applying an action or context transition,
we introduce a special marker $\midState$, which is an ABox assertion with a
fresh concept name $\midStateConcept$ and a fresh constant
$\midStateIndividuals$. When $\midState$ is present, it means that the state
has been produced by an action execution, and that the next transition will
represent a context change.  Such states can be considered as
intermediate, in the sense that the overall change both of the ABox facts and of
the context has not taken place yet.

Formally, given a \cskab $\cskabSym=\cskabTup$, the context-sensitive
transition system $\TS{\cskabSym} =\TSTup$ is defined as follows:
\begin{compactitem}
\item $s_0 = \tup{id_0,A_0,\emptyset,C_0}$;
\item $\stateSet$ and $\trans$ are defined by simultaneous induction as the
  smallest sets satisfying the following properties:
  \myi $s_0 \in \stateSet$;
  \myii if $\tup{id,A,\scmap,C}\in\stateSet$ and $\midState\notin A$, then for
    all actions $\act \in \actSet$,
    for all substitutions $\sigma$ for the parameters of $\act$, and
    for all $\A'$, $\scmap'$ s.t.\
    $\tup{\tup{\A,\scmap,C},\act\sigma,\tup{\A',\scmap',C}} \in
    \exec{\cskabSym}$, let
    \[
      S=\{\tup{id'',A',\scmap',C'}\mid
      \begin{array}[t]{@{}l}
        id'' \text{ is a fresh identifier, and there is } \tup{A',\scmap',C}\\
        \text{such that } \tup{\tup{A',\scmap',C},\tup{A',\scmap',C'}} \in
        \cexec{\cskabSym} \}.
      \end{array}
    \]
    If for some $\tup{id'',A',\scmap',C'}\in S$, we have that $A'$ is
    $\cTBoxSym^{\ctx'}$-consistent, then $s'\in\stateSet$ and
    $\tup{id,A,\scmap,C}\trans s'$,
    where $s'=\tup{id',A'\cup\{\midState\},\scmap',C}$ and $id'$ is a fresh
    identifier.  Moreover, in this case, for each
    $s''=\tup{id'',A',\scmap',C'}\in S$ such that $A'$ is
    $\cTBoxSym^{\ctx'}$-consistent, we have that $s''\in\stateSet$ and
    $s'\trans s''$.
\end{compactitem}
Notice that, if at some point in the above inductive construction, for no
$\tup{id'',A',\scmap',C'}\in S$ we have that $A'$ is
$\cTBoxSym^{\ctx'}$-consistent, then neither the state $s'$ nor any state in
$S$ becomes part of $\stateSet$.

%\endinput

%%% Local Variables:
%%% mode: latex
%%% TeX-master: "main"
%%% save-place: t
%%% End:

%\input{5-CSKAB-Verification}

\section{Verifying Temporal Properties over \cskab}
\label{sec:CSKABVerification}

Given a \cskab $\cskabSym$, we are interested in verifying whether the
evolution of $\cskabSym$, which is represented by $\TS{\cskabSym}$, complies
with some given temporal property.  The challenge is that in general the
transition system is infinite due to the presence of services calls, which can
introduce arbitrary fresh values into the system.

\subsection{Verification Formalism: Context-Sensitive FO-variant of
 $\mu$-Calculus}
\label{sec:VerificationFormalism}

In order to specify temporal properties over \cskabs, we use a first-order
variant of $\mu$-calculus \cite{Stir01,Park76}, one of the most powerful
temporal logics, which subsumes LTL, PSL, and CTL* \cite{ClGP99}. In
particular, we introduce the language \mulcs of \emph{context-sensitive
 temporal properties}, which is based on \muladom defined in
\cite{BCMD*13}.
Basically, we exploit ECQs to query the states, and support a first-order
quantification across states, where the quantification ranges over the
constants in the current active domain.  Additionally, we augment ECQs with
context expressions, which allows us to check also context information while
querying states.
Formally, \mulcs is defined as follows:
\[
  \Phi ~:=~ Q ~\mid~ \ctxl ~\mid~ \lnot \Phi ~\mid~ \Phi_1 \land \Phi_2
  ~\mid~ \exists x.\Phi ~\mid~ \DIAM{\BOX{\Phi}} ~\mid~ \BOX{\BOX{\Phi}}
  ~\mid~ Z ~\mid~ \mu Z.\Phi
\]
where
\begin{inparablank}
\item $Q$ is a possibly open EQL query that can make use of the distinguished
  constants in $\adom{\initABox}$,
\item $\ctxl$ is a context expression over $\langCtx$, and
\item $Z$ is a second order predicate variable (of arity 0).
\end{inparablank}
We adopt the usual abbreviations of FOL, and also
\begin{comment}
$\forall x.\Phi =\lnot(\exists x.\lnot\Phi)$,
$\Phi_1\lor\Phi_2=\lnot(\lnot\Phi_1\land\lnot\Phi_2)$,
\end{comment}
$\BOX{\Phi}=\lnot\DIAM{\lnot\Phi}$
and $\nu Z.\Phi =\lnot\mu Z.\lnot\Phi[Z/\neg Z]$.
Hence $\DIAM{\DIAM{\Phi}}=\lnot\BOX{\BOX{\lnot\Phi}}$ and
$\BOX{\DIAM{\Phi}}=\lnot\DIAM{\BOX{\lnot\Phi}}$.

Notice that $\DIAM{\BOX{\Phi}}$ and $\BOX{\BOX{\Phi}}$ are used in \mulcs to
quantify over the successor states of the current state, obtained after a
state-changing transition followed by a context-changing one.  This allows one
to separately control how the property quantifies over state and context
changes.  Furthermore, due to the fact that the diamond and box operators can
be only used in pairs, the local queries that inspect the data and the context
maintained by the states are never issued over intermediate states, but only
over those resulting from the combination of an action and context transition.

The semantics of \mulcs is defined over a transition system
\mbox{$\TS{}=\TSTup$}.  Since \mulcs contains formulae with both individual and
predicate free variables, given a transition system $\TS{}$, we introduce an
individual variable valuation $\vfo$, i.e., a mapping from individual variables
$x$ to $\const{}$, and a predicate variable valuation $\vso$, i.e., a mapping
from predicate variables $Z$ to subsets of $\stateSet$.  The semantics of
\mulcs follows the standard $\mu$-calculus semantics, except for the semantics
of queries and of quantification. We assign meaning to \mulcs formulas by
associating to $\TS{}$ and $\vso$ an \emph{extension function} $\MODA{\cdot}$,
which maps \mulcs formulas to subsets of $\stateSet$.
The extension function $\MODA{\cdot}$ is defined inductively as follows:
\[
\small
  \begin{array}{r@{\ }l@{\ }l@{\ }l}
    \MODA{Q} &=&
     \{s\in\stateSet \mid \Ans{Q\vfo,\cTBoxSym^{\ctx},\abox(s)}=\mathit{true}\}\\
    \MODA{\ctxl} &=&\{s\in\stateSet \mid \cntx(s)\cup\ctxTh \models \ctxl\}\\
    \MODA{\exists x. \Phi} &=&\{s\in\stateSet \mid\exists d.d\in\adom{\abox(s)}
      \text{ and } s \in \MODAX{\Phi}{[x/d]}\}\\
    \MODA{Z}  &=& V(Z) \subseteq  \stateSet\\
    \MODA{\lnot \Phi} &=& \stateSet - \MODA{\Phi}\\
    %\MODA{\Phi_1 \land \Phi_2} &=& \MODA{\Phi_1}\cap\MODA{\Phi_2}\\
    \MODA{\Phi_1 \lor \Phi_2} &=& \MODA{\Phi_1}\cup\MODA{\Phi_2}\\
    %\MODA{\exists x \in \CONST_0. \Phi} &=& \bigcup\{{\MODA{\Phi}}_{[x/c]}
    % \mid c \in \adom{\mapping(\DBinit)}\}\\
    \MODA{\DIAM{\Phi}}  &=&  \{s\in\stateSet \mid\exists s'.\
      s \Rightarrow s' \text{ and } s'\in\MODA{\Phi}\}\\
    % \MODA{\BOX{\Phi}}  &=&  \{\tup{id, \ABox} \in  \TSSs \mid\forall
    % \tup{id', \ABox'}.\
    % \tup{id, \ABox} \TSSt \tup{id', \ABox'} \\
    % & &\mbox{ implies } \tup{id', \ABox'} \in \MODA{\Phi}\}\\
    \MODA{\mu Z.\Phi} &=&
      \bigcap\{\E\subseteq\stateSet \mid {\MODA{\Phi}}_{[Z/\E]} \subseteq\E \}
    % \MODA{\nu Z.\Phi} & = &
    % \bigcup\{ \E\subseteq  \TSSs \mid  \E \subseteq
    % \MODAX{\Phi}{[Z/\E]}\}
  \end{array}
%\vspace*{-1mm}
\]
%}
where $Q\vfo$ is the query obtained from $Q$ by substituting its free variables
according to $\vfo$.
For a closed formula $\Phi$ (for which $\MODA{\Phi}$ does not depend on $\vfo$
or $\vso$), we denote with $\MOD{\Phi}$ the extension of $\Phi$ in $\TS{}$, and
we say that $\Phi$ holds in a state $s\in\stateSet$ if $s\in\MOD{\Phi}$.
% In this case, we write $\mf{},s \models \Phi$. A closed formula $\Phi$ holds
% in $\mf{}$, denoted by $\mf{} \models \Phi$, if $\mf{},s_0\models \Phi$.

\emph{Model checking} is the problem of checking whether $s_0\in\MOD{\Phi}$,
denoted by $\TS{}\models\Phi$.
We are interested in \emph{verification} of \mulcs properties over \cskabs,
i.e., given a \cskab $\cskabSym$, and a \mulcs property $\Phi$, check whether
$\TS{\cskabSym}\models \Phi$.

\begin{example}
\small
In our running example, the property $\nu Z.(\forall x. \exo{CustOrder}(x)
\wedge \cval{\exc{S}}{\exv{PS}} \limp \mu Y.(\exo{Delivered}(x) \lor
\BOX{\BOX{Y}})) \land \BOX{\BOX{Z}}$ checks that every customer order placed
during peak season will be eventually delivered, independently on how the
context and the state evolve.
\end{example}

\subsection{Decidability of Verification}

In general, verification of temporal properties over \cskabs is undecidable,
even for properties as simple as reachability, which can be expressed in
much weaker languages than \mulcs. This follows immediately
from the fact that \cskabs generalize \kabs \cite{BCMD*13}.

In order to establish decidability of verification, we need to pose
restrictions on the form of \cskabs.  We adopt the semantic restriction of
\emph{run-boundedness} identified in \cite{BCMD*13}, which intuitively
imposes that along every run the number of distinct values cumulatively
appearing in the ABoxes of the states in the run is bounded.  Formally, given a
\cskab $\cskabSym$, a run $\tau = s_0s_1\cdots$ of $\TS{\cskabSym}$ is
\emph{bounded} if there exists a finite bound \emph{b} s.t.\ $\left|\bigcup_{s
   \text{ state of } \tau } \adom{\abox(s)} \right| < b$. We say that
$\cskabSym$ is \emph{run-bounded} if there exists a bound \emph{b} s.t. every
run $\tau$ in $\TS{\cskabSym}$ is bounded by \emph{b}. The following
result shows that the decidability of verification for run-bounded
\kabs can be lifted to \cskabs as well.

\begin{theorem}
  \label{thm:dec}
  Verification of \mulcs properties over run-bounded \cskabs is decidable, and
  can be reduced to finite-state model checking.
\end{theorem}

%-------- Weakly Acyclic --------

\asa{\begin{theorem}
\label{thm:wa}
Given a weakly acyclic \cskab $\cskabSym$, we have that
$\TS{\cskabSym}$ is run-bounded.
\end{theorem}}
%
%\asa{
From Theorems \ref{thm:dec} and \ref{thm:wa}, we finally obtain:
%}
%
%\asa{
\begin{corollary}
\label{thm:decWA}
Verification of \mulcs properties over weakly acyclic \cskabs is
decidable, and can be reduced to finite-state model checking.
\end{corollary}
%}

%\endinput

%%% Local Variables:
%%% mode: latex
%%% TeX-master: "main"
%%% save-place: t
%%% End:

%\input{6-conclusion}

\section{Conclusion}
\label{sec:Conclusion}

We have introduced context-sensitive KABs, which extend KABs with contextual
information.  In this enriched setting, we make use of context-sensitive
temporal properties based on a FOL variant of $\mu$-calculus, and establish
decidability of verification for such logic over \cskabs in which the data
values encountered along each run are bounded.

In this work, we adopt a simplistic approach to deal with inconsistency, based
on simply rejecting inconsistent states. This approach is particularly critical
in the presence of contextual information, which could lead to an inconsistent
state simply due to a context change. In this light, it is particularly
interesting to merge the approach presented here with the one in
\cite{CKMSZ13}, where inconsistency is treated in a more
sophisticated way.

%\endinput

%%% Local Variables:
%%% mode: latex
%%% TeX-master: "main"
%%% save-place: t
%%% End:

\smallskip
\noindent
\textbf{Acknowledgments.}
This research has been partially supported by the EU IP project Optique
(\emph{Scalable End-user Access to Big Data}), grant agreement n.~FP7-318338,
and by DFG within the Research Training Group ``RoSI'' (GRK~1907).

\bibliographystyle{splncs03}
\bibliography{main-bib}

\end{document}